\documentclass[letterpaper, 10 pt, conference]{ieeeconf}  

\IEEEoverridecommandlockouts                              

\usepackage{graphics} 
\usepackage{graphicx} 
\graphicspath{ {Figures/}}
\usepackage{epsfig} 
\usepackage{mathptmx} 
\usepackage{times} 
\usepackage{amsmath} 
\usepackage{amssymb}  
\usepackage{mathtools}
\usepackage{booktabs}
\usepackage{wasysym}
\usepackage{siunitx}
\usepackage{morefloats}
\usepackage[noadjust]{cite}

\usepackage{color,soul}
\usepackage{textcomp}
\usepackage{gensymb}
\usepackage{array}
\usepackage{makecell}
\usepackage{tabularx}
\newcolumntype{K}[1]{>{\centering\arraybackslash}p{#1}}
\usepackage{tabulary}
\usepackage{float}
\usepackage{notoccite}
\usepackage[ruled,vlined]{algorithm2e}
\SetKwInput{KwInput}{Input}
\SetKwInput{KwOutput}{Output}
\usepackage{threeparttable}
\usepackage{nicefrac}
\usepackage[hidelinks]{hyperref}
\usepackage[none]{hyphenat}

\usepackage[table]{xcolor}
\definecolor{blue_background}{HTML}{E3E4FA}
\definecolor{blue_2}{HTML}{E3F2FD}

\title{\LARGE \bf
Switch-based Independent Antagonist Actuation with a Single Motor for a Soft Exosuit
}

\author{Atharva Vadeyar, Rejin John Varghese, Etienne Burdet and Dario Farina
\thanks{This research is supported in part by the UK EPSRC EP/T020970/1 NISNEM and by the EU H2020 REHYB (ICT 871767) grants.} 
\thanks{All authors are with the Department of Bioengineering, Imperial College of Science, Technology and Medicine, London W12 0BZ, UK (email: 
 {\tt\footnotesize \{r.varghese15,e.burdet,d.farina\}@imperial.ac.uk})
 }
}

\begin{document}

\maketitle
\thispagestyle{empty}
\pagestyle{empty}

\begin{abstract}

The use of a cable-driven soft exosuit poses challenges with regards to the mechanical design of the actuation system, particularly when used for actuation along multiple degrees of freedom (DoF). The simplest general solution requires the use of two actuators to be capable of  inducing movement along one DoF. However, this solution is not practical for the development of multi-joint exosuits. Reducing the number of actuators is a critical need in multi-DoF exosuits. We propose a switch-based mechanism to control an antagonist pair of cables such that it can actuate along any cable path geometry. The results showed that 298.24ms was needed for switching between cables. While this latency is relatively large, it can reduced in the future by a better choice of the motor used for actuation.

\indent \textit{Clinical relevance}— This work proposes a simplified mechanism to generalise the design of soft exosuits and exo-gloves. The mechanism will result in a generalisable exoskeleton platform for at-home and hospital-based rehab and assistive applications. 
\end{abstract}

\section{Introduction} 
Conditions affecting motor functions may be due to multiple factors such as age, injuries, or diseases such as Amyotrophic Lateral Sclerosis (ALS) or Parkinson’s. The use of exoskeletons to manage physical manifestations of these conditions is being extensively studied. While rigid exoskeletons find applications predominantly requiring larger assistance where weight and cost is not a prohibitory factor, soft systems can become more ubiquitous in applications for partial assistance \cite{5-xiloyannis2021soft,4-varghese2018wearable}. ‘Soft’ exoskeletons function as an external layer of muscle, relying structurally on the human body to transfer assistive forces \cite{5-xiloyannis2021soft}. The materials used in these devices allow for greater portability, comfort, and affordability, making them accessible to a broader population \cite{4-varghese2018wearable}. 

Single degree-of-freedom (DoF) soft exoskeletons capable of assisting simple or task-specific movements such as flexion/extension are limited in their ability to assist with more general activities of daily living (ADL). ADL typically require simultaneous use of multiple 1-DoF and multi-DoF joints. Amongst soft exosuits, cable-driven systems powered by motors are most prevalent as this actuation technology is mature, provides favourable power density, and with a Bowden-cable arrangement allows for the joint being assisted to be relatively unencumbered as the heavier elements of the exoskeleton, such as the actuation unit, can be placed remotely. 

\begin{figure}
    \centering
    \includegraphics[width=0.85\linewidth]{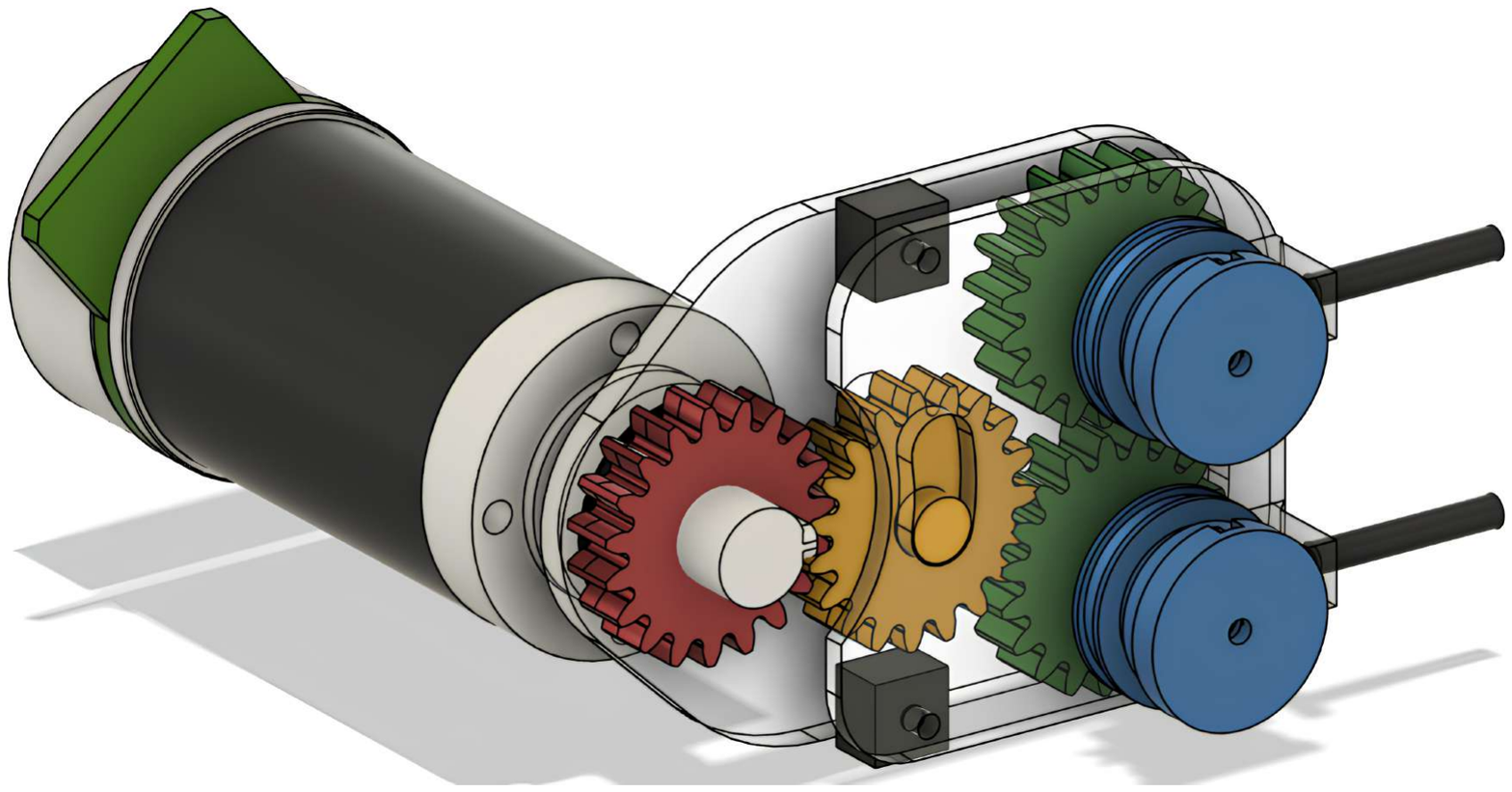}
    \caption{The proposed actuation mechanism capable of independently facilitating antagonist actuation.}
    \label{fig:fig1}
\end{figure}

When two cables are arranged in an antagonist configuration to assist a 1-DoF joint, they have geometrically different cable paths and, subsequently, cable displacements. This introduces a non-linear dependence between these cable lengths. The exact nature of this dependence may vary based on the user and the routing of the cables through the system. The simplest solution is to actuate each cable with a separate motor, requiring the use of 2 motors for each joint/DoF. However, this can lead to the overall system becoming heavy and expensive, diminishing the inherent benefits of softer systems. 

To solve this challenge, several strategies have been proposed such as underactuation based on desired function \textit{e.g.} grasping\cite{13-2015tendon_slack_feasibility,14-2016tendon_slack} or, more broadly, synergies \cite{11-xiloyannis2016modelling, 12-alicea2021soft}, one-to-many actuation (OTM) mechanisms \cite{29-xiloyannis2019design}, differential drives \cite{18-bajaj2020soft}, among others. Function and synergy-based underactuation mechanisms require specifically tuned pulley radii or cam profiles which needs to be individual-specific, a solution that is not generalisable. OTM, while potentially more generalisable, end up weighing about the same as using individual motors, suffer from delays and a lack of smoothness. Differential drives and similar solutions again end up being complex and bulky and unsustainable for a wearable robot. These challenges pose the need for a low-complexity device that can serve as a more general solution.

\begin{figure*}
    \centering
    \includegraphics[width=0.85\linewidth]{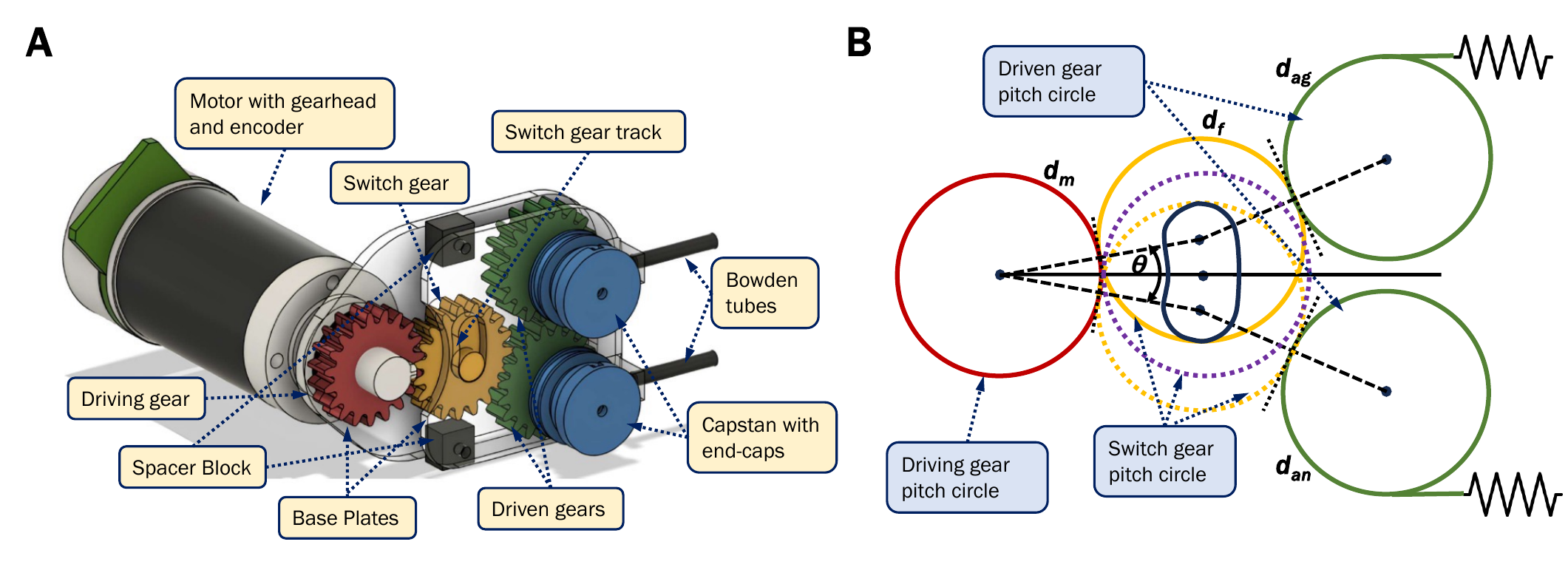}
    \caption{Schematic of the switch-based actuation mechanism: A). CAD rendering of the mechanism with labelled components, B). Geometrical configuration of the pitch circles of different gears and their placement. It is to be noted that the driven gears are spring-loaded.}
    \label{fig:fig2}
\end{figure*}

In this paper, we propose a switch-based mechanism capable of using a single motor to independently drive antagonist motion of one DoF, using a switch-based mechanism (Fig.\ref{fig:fig1}). The solution was devised to be of low complexity, affordable and lightweight. Through the paper we first discuss the concept of the mechanism, followed by experimental studies to understand its capability in ensuring independence of both the agonist and antagonist movement as well as reasonable delays in switching between either direction of movement, concluding with a discussion on the features, limitations and future improvements. 

\section{The Switch-based Actuation Mechanism}

As mentioned previously, in cable-driven systems the antagonist paths generally have geometrically different displacements. Analogous to hard-coding, the simplest 1-motor solution would be to customise the pulley profiles of both cables \cite{31-TaskDefinedPulley}. However, this results in a bespoke and non-generalisable solution. To address this issue, we aimed for our mechanism to achieve antagonist motion independently. Consequently, the proposed design was developed to allow the antagonist (unwinding spool) to be decoupled from the agonist (winding spool), in order to allow for the unwinding spool to rotate as needed to cover a given unwinding path. 

Figs.\ref{fig:fig1}\&\ref{fig:fig2} show the mechanism for the system. A gear placed within a track functions as the switch. The switch gear (yellow) is always meshed with the motor and forms a truncated planetary gear system, with the motor gear (red) being the ‘Sun’ gear and the switch being the ‘Planet’ gear. When the motor switches direction, the switch gear disengages from its presently engaged driven gear (green) and begins revolving around the motor in the same direction along the track. This happens because the gear is not held fixed in place, and hence, the intrinsic friction between the driving and switch gear causes the switch gear to move in the track. As it travels across the midline (Fig.\ref{fig:fig2}B), it starts engaging with the other driven gear, and at the end of the track begins rotating in place, subsequently rotating the spool. When the switch gear is very close to the centre (purple circle in Fig.\ref{fig:fig2}B), the motor is decoupled from both the spools, which is a safety feature. 

This mechanism ensures that when the agonist cable is pulled, the driven gear connected to the antagonist cable is disengaged allowing for independent lengthening of the antagonist cable. However, to prevent uncontrolled unspooling and slack of the cable, both the spools are spring-loaded with a clock-spring to ensure the disengaged cable is always in tension. As can be inferred from Fig.\ref{fig:fig2}B, depicting the pitch circles of the driving (red), switch (yellow) and driven (green) gears, the angle $\theta$ decides the total travel, and consequently switching delay (along with motor speed). This is a function of the radii of the four gears and the modulus of the gears.

\begin{figure*}
    \centering
    \includegraphics[width= 0.7\linewidth]{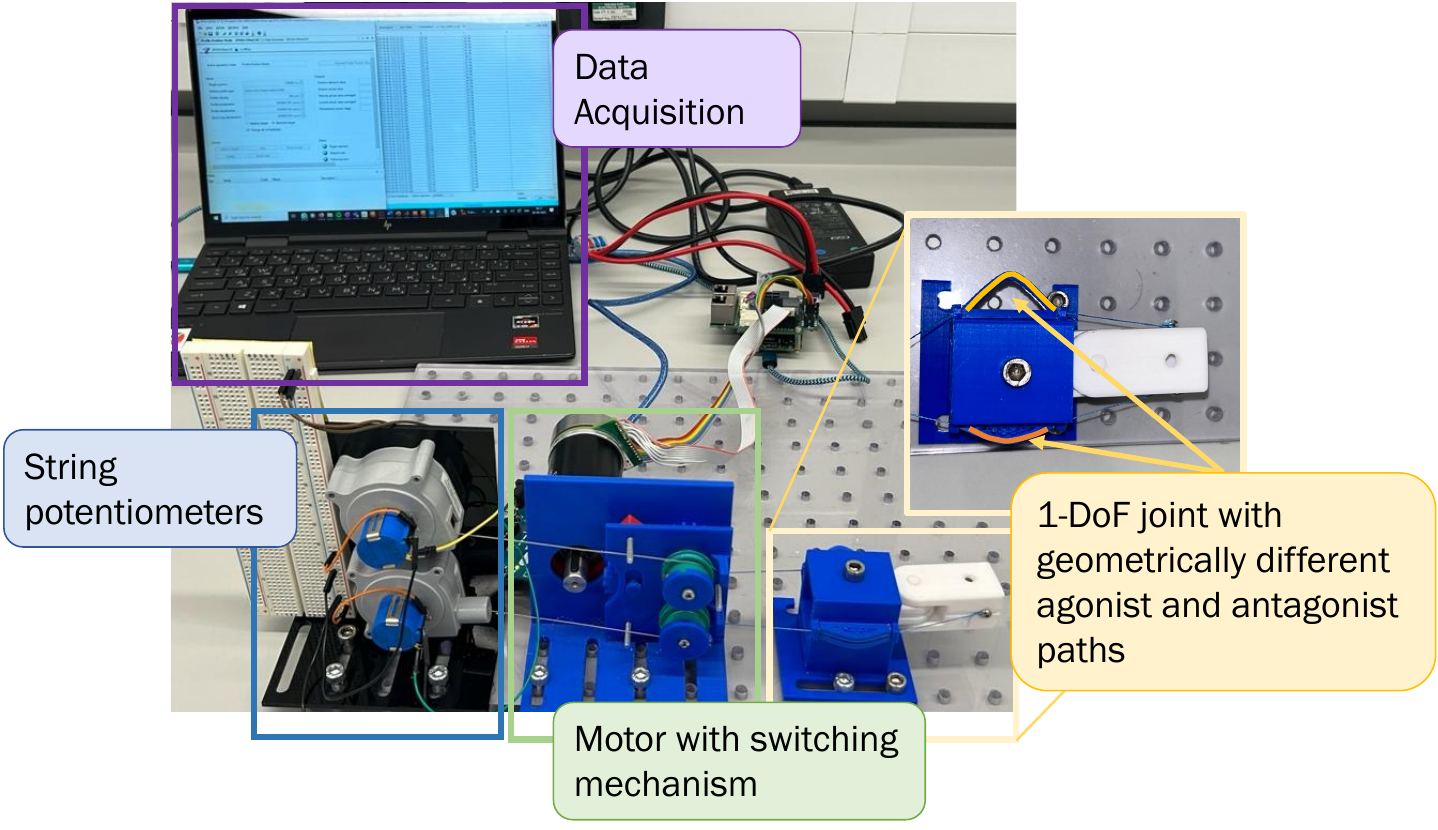}
    \caption{The test rig to test the switching mechanism. It consists of a motor with the switching mechanism. The capstans in the mechanism are connected to both the string potentiometers and the simulated 1-DoF joint with different tendon paths. }
    \label{fig:fig3}
\end{figure*}

\section{Validation Experiments \& Results}

\subsection{Experimental Setup}

A prototype of the concept and a test platform were built for validating the concept (Fig.\ref{fig:fig3}). The design was realised in Fusion360 (Autodesk, CA, USA) and 3D-printed. The weight of the entire mechanism was 56g (excluding motor and string potentiometers), which is light for an actuation unit. While this can be further miniaturised, it should be noted that better materials will be needed for handling higher torques. The switch-mechanism was combined with an EC-45 Flat motor (Maxon, Switzerland) with a 43:1 gearhead and encoder. The reduction results in a maximum motor output speed of 120rpm. An EPOS4 EtherCAT 50/8 controller was used to control the motor. The two capstans/spools were each connected to a Celesco SP1-12 string potentiometer (TE Connectivity, Galway, Ireland), simultaneously, providing the required spring behaviour and displacement measurement. Braided nylon fishing line was used to connect the spools to a hinged joint. The cables were routed via two different curved profiles to simulate the different cable paths observed in soft exoskeletons and terminated in a 1-DoF joint with a range of motion (RoM) of  [$-90\degree, 90\degree$] (Fig.\ref{fig:fig3}). The agonist and antagonist cables would be alternately pulled to achieve full range of motion of the joint. Cable displacement using the string potentiometers was recorded using an Arduino Uno (Arduino, Monza, Italy). The EPOS4 controller was used for position, velocity and current (torque) control and recording data.

\subsection{Experimental Protocol}

To test the proposed concept, it was necessary to ensure that the spools were independent from each other, i.e. when one is actuated the other can freely be pulled under tension. Secondly, even after independence in both degrees of freedom, it is crucial for the switching time to be practical for assisting ADL and rehabilitation-based applications. The experimental protocols for both these tests are described next.

\subsubsection{Spool Independence Test}

The joint was driven over the full RoM ( $[-90\degree,90\degree]$), and while moving in one direction, the unwinding spool was randomly disturbed. The aim of this experiment was to demonstrate that the spooling of the agonist cable does not affect the unspooling of the antagonist cable. It should be noted that the cables for the agonist and antagonist are traversing geometrically different paths during this test (Fig.\ref{fig:fig3}).

\subsubsection{Switching Time Test}
The motor was driven in position control mode (Profile Position Mode in EPOS Studio) between two empirically determined points that correspond to moving from one extreme to another in the track without engaging either driven gear/spool. The test was repeated ten times and the duration for traversal was recorded.

\subsection{Results}

\begin{figure}
    \centering
    \includegraphics[width=0.85\linewidth]{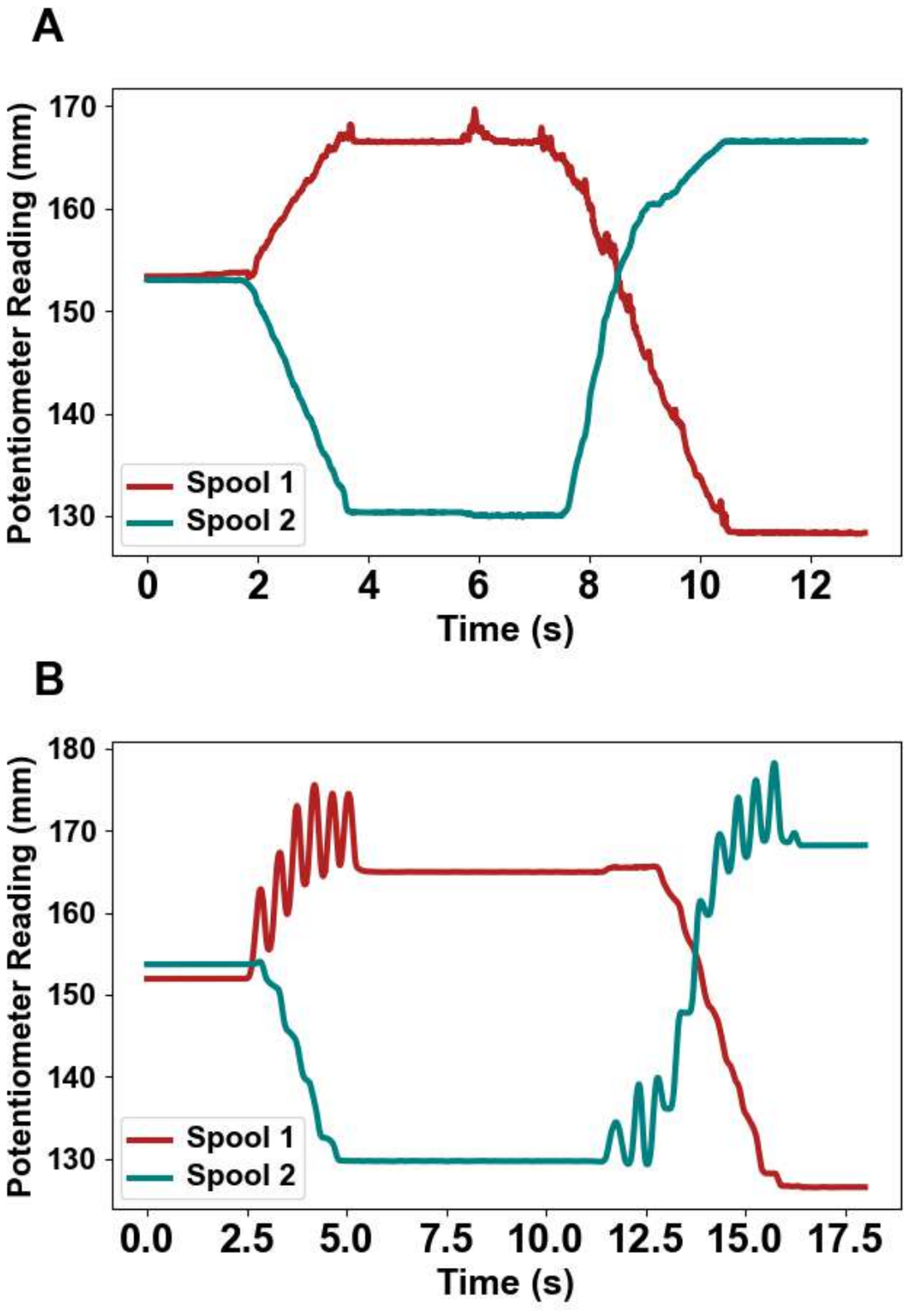}
    \caption{Spool Independence Test Results}
    \label{fig:fig4}
\end{figure}

As mentioned previously, testing independence of the agonist and antagonist spools and the time to switch between either driven spool is vital to understand the practical viability of the concept. During the experimental trial to test spool independence, we could clearly observe the agonist and antagonist cables moving in opposite directions (Fig.\ref{fig:fig4}A) and how inducing disturbances in the unwinding cable (see perturbations in the red and green graphs in Fig.\ref{fig:fig4}B) did not affect the driven cable. This clearly showcases the design concept's ability to accommodate various cable paths, enabling the same actuation unit to be generalisable. 

\begin{figure}
    \centering
    \includegraphics[width=0.85\linewidth]{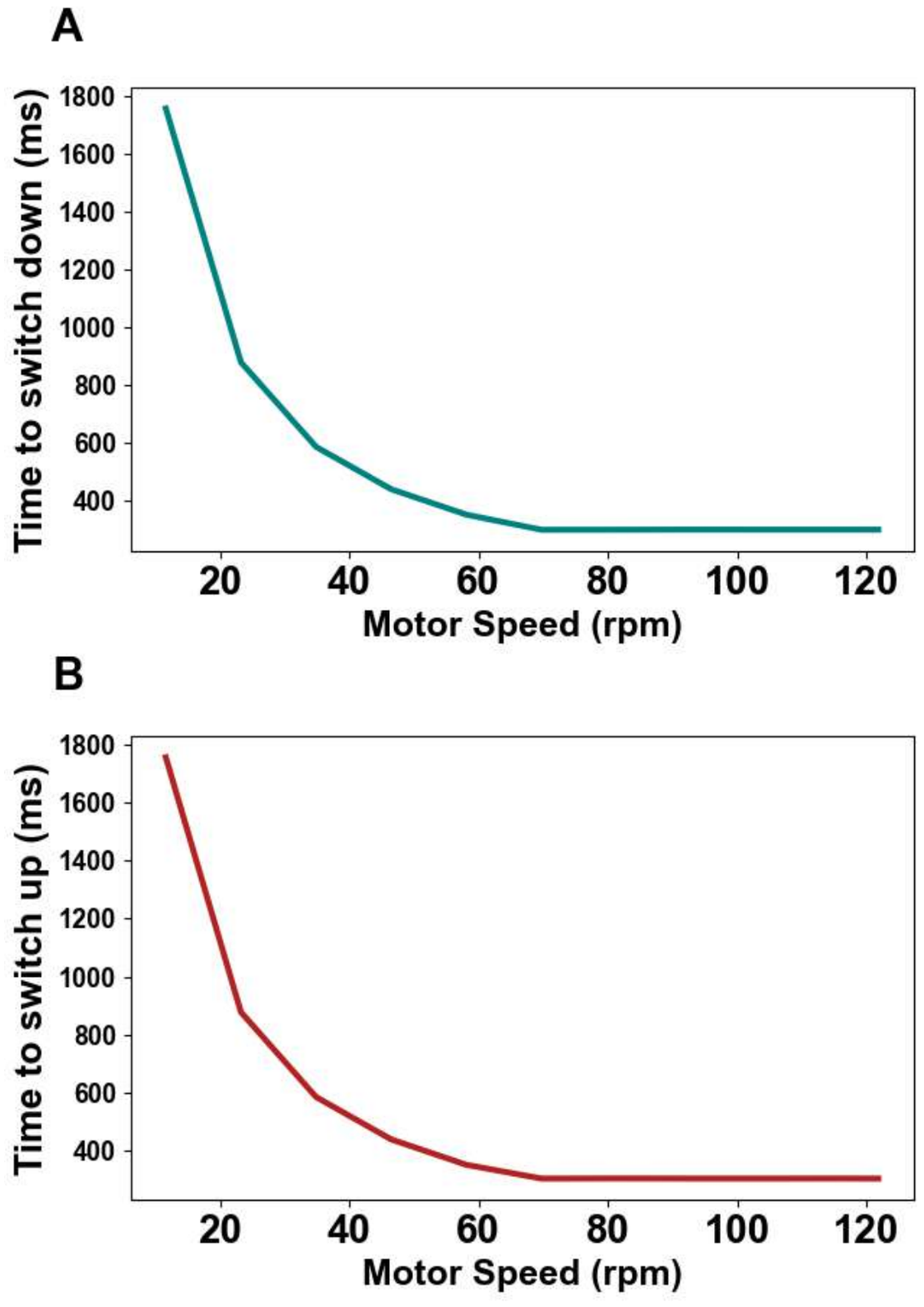}
    \caption{Switching Time Test Results}
    \label{fig:fig5}
\end{figure}

In conducting the switching time tests, average times of 302ms ($\sigma$=0.506) and 298ms ($\sigma$=0.640) were found over 10 trials for the switch to move up and down, respectively, with a motor output speed ($\omega$) of $\approx120$rpm. Given fixed geometry of the gears, the switching time ($\propto\frac{\theta}{\omega}$) is dependent on the motor speed and is presented in Fig.\ref{fig:fig5}. However, it should be noted that a simple adjustment in the gear ratio between the driving and floating gear can help bringing the switching delay down significantly. It is also observed that gravity does not play a role in altering switching times, and the friction between driving and switch gear is sufficient to engage either spool reliably (variability in switching times very low). It is important to note that the motor output shaft rotates $122.6\degree$ to make the switch gear to revolve around the driving/motor gear by $19.8\degree$. This depends on the relative friction between the track and switch gear shaft (friction in revolution) v/s the friction between the driving and switch gear teeth (friction in rotation).

\section{Discussion \& Conclusion}

The proposed switching mechanism could become an adequate replacement to alternatives such as fixed agonist/antagonist pulley profiles (for underactuation or task/synergy-specific movements), OTM actuation mechanisms, differential drives \textit{etc.}, especially in the context of using a single motor to drive both the agonist and antagonist direction of a joint. The simplicity, light weight and affordability of the proposed mechanism are advantages that make it suitable for multi-DoF platforms. Furthermore, the mechanism also has a safety feature which results from moving the switch-gear to the centre, thereby, disengaging the prime mover from both spools and allowing for transparent movement of the assisted joint. The independence of the agonist and antagonist paths ensures safety and avoids any undesirable forces from being applied during assistance. With regards to the switching delay of 300ms for this particular unoptimised choice of motor and gear sizes, we acknowledge this as a limitation for highly dynamic activity assistance. However, for rehabilitation applications and ADL which generally take place in the range of 1-3Hz, this will not be a limitation. As this prototype was developed to validate the concept, we intend to optimise the design of the mechanism to especially minimise the footprint and the switching delay. Another limitation which is not addressed in this mechanism (as well as in the OTM and underactuation strategies that are based on a single actuator/motor), is the inability to provide both torque and stiffness on demand, which is an area of focus for future development. However, we believe this switching mechanism is an apt starting point for further research and development into practical actuators for multi-DoF exosuits.

\section*{Acknowledgement}
This work utilised expertise and prototyping equipment at the Imperial College Advanced Hackspace. The authors would like to thank the mentors at the Hackspace for their support. 


\bibliographystyle{IEEEtran}
\bibliography{IEEEabrv,atharva_references}

\end{document}